# Self-Organized Operational Neural Networks for Severe Image Restoration Problems

Junaid Malik[a], Serkan Kiranyaz[b] and Moncef Gabbouj[a]

[a]Tampere University, Tampere, Finland
[b]Qatar University, Doha, Qatar

*Abstract*— Discriminative learning based on convolutional neural networks (CNNs) aims to perform image restoration by learning from training examples of noisy-clean image pairs. It has become the go-to methodology for tackling image restoration and has outperformed the traditional non-local class of methods. However, the top-performing networks are generally composed of many convolutional layers and hundreds of neurons, with trainable parameters in excess of several millions. We claim that this is due to the inherent linear nature of convolution-based transformation, which is inadequate for handling severe restoration problems. Recently, a non-linear generalization of CNNs, called the operational neural networks (ONN), has been shown to outperform CNN on AWGN denoising. However, its formulation is burdened by a fixed collection of well-known non-linear operators and an exhaustive search to find the best possible configuration for a given architecture, whose efficacy is further limited by a fixed output layer operator assignment. In this study, we leverage the Taylor series-based function approximation to propose a self-organizing variant of ONNs, Self-ONNs, for image restoration, which synthesizes novel nodal transformations on-the-fly as part of the learning process, thus eliminating the need for redundant training runs for operator search. In addition, it enables a finer level of operator heterogeneity by diversifying individual connections of the receptive fields and weights. We perform a series of extensive ablation experiments across three severe image restoration tasks. Even when a strict equivalence of learnable parameters is imposed, Self-ONNs surpass CNNs by a considerable margin across all problems, improving the generalization performance by up to 3 dB in terms of PSNR.

## 1. Introduction

Image restoration aims at recovering low-level contextual information from noisy and corrupted images. It is one of the key inverse imaging computer vision tasks because the quality of image acquisition is inherently subdued by environmental conditions, quality of the image capturing device and processes involved in obtaining a digital image from photo sensors. Earliest works were based on the basic assumption that smoothing by averaging leads to denoising (Alvarez, Lions, & Morel, 1992; Donoho, 1995; Smith & Brady, 1997). Such methods were then surpassed by non-local class of methods (Buades, Coll, & Morel, 2005; Mahmoudi & Sapiro, 2005; Mairal, Bach, Ponce, Sapiro, & Zisserman, 2009), among which the method BM3D (Dabov, Foi, Katkovnik, & Egiazarian, 2007) is widely considered to be state-of-the-art. In recent years, the focus has shifted towards *learning* the restoration problem. by posing it in a discriminative paradigm of training on noisy-clean image pairs. Convolutional Neural Networks (CNNs)-based approaches have rapidly reached apex performance in almost all learning-based computer vision problems (He, Zhang, Ren, & Sun, n.d.; Krizhevsky, Sutskever, & Hinton, 2012; Shelhamer, Long, & Darrell, 2017), and image restoration is no exception (Lempitsky, Vedaldi, & Ulyanov, 2018; Zhang, Zuo, Chen, Meng, & Zhang, 2017; Zhang, Zuo, Gu, & Zhang, 2017).

CNNs are artificial neural networks composed of stacked layers of convolutional neurons, each filtering local receptive fields within the input feature maps by convolving them with learnable filter banks. Their weight sharing and local connections significantly reduce the size of parameter space as compared to multilayer perceptrons (MLPs) and makes them especially efficient for large grid-structured data such as images. While the earlier CNN models were not particularly deep (LeCun, Bottou, Bengio, & Haffner, 1998), faster implementations on GPU (Cires, Meier, Masci, & Gambardella, 2003) ushered in an era of deeper and more complex architectures [6],[7]. Contemporary CNN architectures are generally composed of tens of layers and their performance is generally observed to be correlated with their *depth* (number of layers) (Conneau, Schwenk, Cun, & Barrault, 2017). Some of the state-of-the-art CNNs for image restoration consist of learnable parameters in the order of millions (Zou, Lan, Zhong, Liu, & Luo, 2019), despite dealing with mild restoration problems where the corrupted images still retain most of the semantic information.

Despite their wide-scale adoption, as identified in (Kiranyaz, Ince, Iosifidis, & Gabbouj, 2020), the necessity of deeper and wider CNN architectures stems from some of the inherent drawbacks in the convolutional model. Firstly, the convolutional neuron model implies a strict linear transformation, where the only source of non-linearity may stem from the point-wise non-linear activation. Therefore, a very high number of neurons with interwoven non-linear activations are required in order to synthesize a rich enough hypothesis for challenging restoration problems. A remedy to this has been recently proposed in (Kiranyaz et al., 2020) using the so-called Operational Neural Networks (ONNs) which embed non-linear operations inside the patch-wise transformations as an alternative to the linear convolutional model. Specifically, the linear transformation of weight multiplications and additions (forming the convolution operation) are generalized to non-linear mappings, called nodal and pool functions, respectively. These nonlinear mappings are defined in an operator set library and searched for each learning problem individually. Similar to their predecessors, Generalized Operational Perceptrons (GOPs), which were shown to be superior to traditional Multi-Layer Perceptrons (MLPs) (Kiranyaz, Ince, Iosifidis, & Gabbouj, 2017a, 2017b; Tran, Kiranyaz, Gabbouj, & Iosifidis, 2020), ONNs were shown to outperform equivalent and even deeper CNN architectures across a variety of computer vision problems, including AWGN image denoising (Kiranyaz et al., 2020).

In ONNs, the choices for nodal and pool operators are critical towards generating an optimal degree of non-linearity for a given problem. While their generic formulation enables the flexibility to incorporate any non-linear transformation, their efficacy relies on two key factors; i) curation of a diverse enough operator set library and ii) the selection of the optimal operators for the problem at hand. In case of the former, it is possible that the optimal non-linear transformation required for the restoration problem cannot be expressed by a well-known function such as a sinusoid or an exponential function. Therefore, there always stands a possibility that the operator set library, no matter how large it is, remains insufficient. Secondly, the search paradigm used in (Kiranyaz et al., 2020) is called Greedy Iterative Search (GIS) that requires a large number of prior training runs for each learning problem, in order to converge towards an optimal set of operator per hidden layer. Therefore, using ONNs for the image restoration problem would imply selecting an appropriate operator set library and searching over it separately for each of the noise characteristics being studied. This makes their usage cumbersome and perhaps impractical for large datasets. Moreover, even if an ideal convergence is assumed, GIS-based operator search only yields a homogenous configuration of layers i.e. all neurons in a given layer are limited to having the same non-linear transformation. Extending it for a heterogenous configuration, where each neuron has a distinct operator set, would make the search prohibitively expensive.

Based on the above facts, in order to solve severe and diverse restoration problems, we identify the need for a self-reliant alternative to convolutional neurons for embedding non-linearities without the need for additional training runs, as in ONNs. To accomplish this objective in this study, we propose, Self-ONN, a self-organized variant to the ONN which does not require prior curation of an operator set library and consequently, the need to search over it. Self-ONNs with generative neurons can leverage the Taylor series-based approximation to generate "any" nodal operator, which is optimized as part of the learning process, thus voiding the need of any prior training runs. In this study, we compare the proposed approach with equivalent CNN and ONN architectures having the same number of learning units (neurons) and network parameters. The experiments are conducted on severely corrupted images from three benchmark datasets and different noise types. An extensive set of experimental results demonstrate that the proposed Self-ONNs exhibit a superior generalization and training performance compared to both equivalent ONNs and CNNs across all image restoration problems.

The rest of the paper is structured as follows. In Section 2, we briefly describe some of the contemporary methods employed for image restoration. Section 3 provides technical details of Self-ONNs by comparing their formulation to vanilla ONNs and CNNs. Section 4 elucidates the experimental setup and the characteristics of networks and datasets used in this study. Section 5 provides key insights into the results and findings of the study, while Section 6 concludes the paper and suggests possible future research directions.

**2. Related Works**

Prior to the advent of CNN-based discriminative learning, the non-local class of methods dominated the field of image restoration (Buades et al., 2005; Mahmoudi & Sapiro, 2005; Mairal et al., 2009). Arguably, the most successful technique belonging to this class is that of BM3D (Dabov et al., 2007). The method works by creating a pool of similar non-local patches across the image. A collaborative filtering procedure is subsequently applied where the pools of patches are projected to a higher-dimensional space and denoised by shrinking the transform coefficients. Recently, CNNs have become the *de facto* standard. Earlier, in (Burger, Schuler, & Harmeling, 2012), it was observed that an MLP can be trained to competitively reproduce the results of BM3D. In (Zhang, Zuo, Chen, et al., 2017), a deep CNN architecture was proposed that successfully applied batch normalization and residual learning principles to achieve competitive results for various degrees of AWGN. The study in (Zhang, Zuo, & Zhang, 2018) aimed to incorporate spatial-invariance by proposing to supplement the inputs with an additional noise map, so that the CNN can learn spatially-invariant encodings for denoising. In (Zhang, Zuo, Gu, et al., 2017), the residual framework adopted in (Zhang, Zuo, Chen, et al., 2017) is used and extended with batch renormalization and dilated convolutions to address the problems with small mini-batch and limited receptive fields, respectively. The authors of (Zou et al., 2019) employed a very deep network, composed of 52 layers, with global and local residual framework to tackle high-level AWGN denoising. Generally, the proposed CNN-based methods employ considerably deep architectures and learn from large-scale datasets; consisting of training examples in the order of $10^5$. Moreover, the noise characteristics of input images are generally mild and preserve the contextual information of the image quite well. Therefore, there is a need to evaluate learning models for restoring highly corrupted images having diverse and severe noise characteristics. Furthermore, as of now, there exists scarce literature that explores the possibility of exploiting non-linearity in the realm of CNNs for image denoising (Kiranyaz et al., 2020; Wang, Yang, Xie, & Yuan, n.d.; Zoumpourlis, Doumanoglou, Vretos, & Daras, 2017). While ONNs provide a viable alternative, the need of defining a fixed operator set library in advance, and performing an exhaustive search to find best operators may render their usage impractical, especially for large-scale datasets.

## 3. Self-Organized Operational Neural Networks

In ONNs, the primary building block is the operational neuron model which extends the principles of GOPs to convolutional realm. While retaining the favorable characteristics of sparse-connectivity and weight-sharing in a CNN, an ONN provides the flexibility to incorporate *any* non-linear transformation within local receptive fields without the overhead of additional trainable parameters. In the following section, we provide a brief overview of the convolutional operation in CNN and explain how ONNs generalize it.

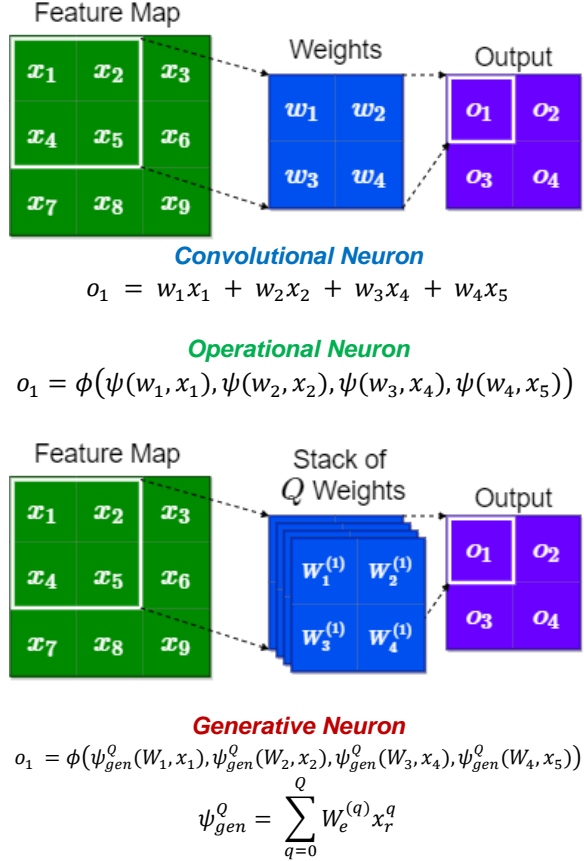

Figure 1 Illustration of the formulations of convolutional, operational and self-organized (generative) operational neuron.

### 3.1. Operational Neural Networks

In a convolutional neuron, given the output of layer $l-1$, the pre-activation output of the $k^{th}$ convolutional neuron in layer $l$ is calculated as:

$$x_l^k(i,j) = \sum_{u=0}^{m-1} \sum_{v=0}^{n-1} w_l^k(u,v) y_{l-1}(i-u, j-v) \quad (1)$$

where $y_{l-1} \in \mathbb{R}^{M \times N}$ and the weight kernel $w_l^k \in \mathbb{R}^{m \times n}$. For the sake of brevity, unit stride and dilation are assumed, and the input is padded with zeros before the convolution operation in order to preserve the spatial dimensions. An alternate formulation of the operation of (1) is now presented. Firstly, $y$ is reshuffled such that values inside each $m \times n$ sliding block of $y_{l-1}$ are vectorized and concatenated as rows to form a matrix $Y_{l-1} \in \mathbb{R}^{\widehat{M} \times \widehat{N}}$ where $\widehat{M} = MN$ and $\widehat{N} = mn$. This operation is commonly referred to as "*im2col*" and is critical in conventional GEMM-based convolution implementations (Chetlur et al., 2014). Secondly, we construct a matrix $W_l^k \in \mathbb{R}^{\widehat{M} \times \widehat{N}}$ whose rows are repeated copies of $\overrightarrow{w_l^k} = vec(W_l^k) \in \mathbb{R}^{mn}$, where $vec(\cdot)$ is the vectorization operator. Each element of $W_l^k$ is given by the following equation:

$$W_l^k(i,j) = \overrightarrow{w_l^k}(i) \qquad (2)$$

The convolution operation in (1) can then be represented as follows:

$$x_l^k = vec_{M \times N}^{-1}\left(\sum_j (Y_{l-1} \otimes W_l^k)\right) \qquad (3)$$

where $\otimes$ represents the Hadamard product, $\sum_j$ is the summation across $j^{th}$ dimension. In (3), $vec_{M \times N}^{-1}$ is the inverse vectorization operation that reshapes back to $M \times N$. The formulation given in (3) can now be generically reformulated to represent the forward-propagation through an operational neuron:

$$x_l^k = vec_{M \times N}^{-1}(\phi_l^k(\psi_l^k(Y_{l-1}, W_l^k))) \qquad (4)$$

where $\psi(\cdot): \mathbb{R}^{M \times N} \to \mathbb{R}^{M \times N}$ and $\phi(\cdot): \mathbb{R}^{\widehat{M} \times \widehat{N}} \to \mathbb{R}^{\widehat{M}}$ are termed as *nodal* and *pool* functions, respectively. Finally, after applying the activation function $f_l^k$, we get the output of the neuron:

$$y_l^k = f_l^k\left(vec_{M \times N}^{-1}\left(\phi_l^k\left(\psi_l^k(Y_{l-1}, W_l^k)\right)\right)\right) \qquad (5)$$

Given an operator set; a triplet of $(\psi_l^k, \phi_l^k, f_l^k)$, an operational neuron implements the formulation given in (5). It is worth noticing here that the convolutional neuron is a special case of an operational neuron with nodal function $\psi(\alpha, \beta) = \alpha * \beta$ and pooling function $\phi(\cdot) = \sum_i$.

*3.2. Self-Organized Operational Neural Networks*

The Taylor series expansion of an infinitely differentiable function $f(x)$ about a point $a$ is given as:

$$f(x) = \sum_{n=0}^{\infty} \frac{f^{(n)}(a)}{n!}(x-a)^n \qquad (6)$$

The $Q^{th}$ order truncated approximation, formally known as the Taylor polynomial, takes the form of the following finite summation:

$$f(x)^{(Q,a)} = \sum_{n=0}^{Q} \frac{f^{(n)}(a)}{n!} x^n \qquad (7)$$

The above formulation can approximate any function $f(x)$ sufficiently well in the close vicinity of $a$. With the use of an activation function which bounds the neuron's input feature maps within $[a - \gamma, a + \gamma]$, the formulation of (7) can be exploited to form a composite nodal operator where the coefficients of the powers of $x$ are the learned parameters of the network. Following the same notation as introduced in Section 3.1, the nodal operator of the $k^{th}$ generative neuron in the $l^{th}$ layer would take the following general form:

$$\psi_l^k(Y_{l-1}, \boldsymbol{W_l^k}, Q, a) = \sum_{q=1}^{Q} Y_{l-1}^q \otimes W_l^{k\,(q)} \qquad (8)$$

where $\boldsymbol{W_l^k} \in \mathbb{R}^{\widehat{M} \times \widehat{N} \times Q}$ is the three-dimensional weight matrix and $W_l^{k(q)} \in \mathbb{R}^{\widehat{M} \times \widehat{N}}$ is the $q^{th}$ slice of $\boldsymbol{W_l^k}$. The $0^{th}$ order term $a$, the DC bias, is ignored as its additive effect can be compensated by the learnable bias parameter of the neuron. Back-propagation (BP) through this nodal operator is now trivial to accomplish. Equations (9) and (10) provide the derivatives with respect to the input $Y_{l-1}$ and the $q^{th}$ slice weights, $W_l^{k\,(q)}$, respectively:

$$\frac{d\psi_l^k}{dY_{l-1}} = \sum_{q=1}^{Q} q Y_{l-1}^{q-1} \otimes W_l^{k\,(q)} \qquad (9)$$

$$\frac{d\psi_l^k}{dW_l^{k\,(q)}} = Y_{l-1}^q \qquad (10)$$

During the learning process, as the weights are updated by BP, this formulation enables the network to spawn novel nodal transformations that are optimized to achieve the given learning objectives. Moreover, the resulting polynomials are functions that provide accurate approximation only within the operating range $[a - \gamma, a + \gamma]$, which makes them easier to compute as opposed to functions that generally comprise the operator set library of an operational neural network (Kiranyaz et al., 2020). In addition, the formulation is heterogenous by design and enables different nodal transformations for different neurons in a layer. The heterogeneity offered by such a formulation is incorporated in all neurons of all the layers, including the output layer. This provides a crucial advantage over ONNs, where the output layer operators are always confined a priori to a fixed operator set (Kiranyaz et al., 2020). Expanding on the intuition that earlier layers of neural networks are involved with feature extraction while the latter ones deal with classification, we can see that self-organizing operational neurons not only can extract more discriminative features, but can also generate better decision boundaries, by virtue of the optimized nodal transformations. Furthermore, as depicted in Figure 1, the set of $Q$-weights corresponding to each element of the $K_x \times K_y$ receptive field are unique, and consequently, correspond to different transformations. This provides an unprecedented level of diversity where all the connections between individual elements of the receptive field to the corresponding weights are governed by different transformation functions (i.e., the nodal operators for ONNs). Such a level of heterogeneity is not possible in the formulation of ONNs since a single nodal operator is assigned to each neuron and thus is used by *all* its connections to the previous layer neurons with distinct weights (parameters). Therefore, the proposed generative neuron is not simply another nodal transformation, but it adds another layer of generalization on top of the operational neuron idea by enabling the fore-mentioned heterogeneity.

Finally, and most importantly, the property of generating nodal transformations on-the-fly alleviates the need for prior training runs to search for an appropriate operator set, which is a practical challenge in case of ONNs. Therefore, an operational neuron equipped with this nodal operator, termed as a *generative neuron*, and the resulting network configuration, the Self-ONN, provides a plug-and-play replacement for the convolution-based models and, therefore, can directly be integrated into any contemporary CNN architecture.

*3.3. Relationship to CNN*

CNNs are a subset of ONNs, corresponding to nodal and pool functions of multiplication and summation, respectively. Similarly, convolution is also a special case of a generative neuron with $Q$ set to 1. Moreover, if the pooling operator $\phi$ is fixed to summation the formulation of (8) can be interpreted as $Q$ independent convolutions. The forward propagation through the neuron can then be achieved using a single large convolution operation. This property enables a fast implementation of generative neurons and provides a significant computational edge when compared to the operational ones.

**4. Experiments**

In order to compare the learning ability of various network architectures under harsh conditions, we apply the following training constraints in our experiments:

i) Training data is kept scarce. Only 10% (100) patches are used to train each model, while the remaining 90% (900) patches, as well as high resolution noisy images from BSD12 and BSD68 datasets, are held out for testing.
ii) Number of epochs is limited to 100.
iii) Model architectures are compact; limited to two hidden layers.
iv) Stochastic Gradient Descent (SGD) with momentum based optimization is used in BP. Adaptive optimizers such as Adam (Kingma & Ba, 2015), and RMSProp (Tieleman & Hinton, 2012), are omitted.
v) In all noise types, the degree of corruption is severe (SNR < 0dB) resulting in the loss of nearly all the contextual information.

These constraints will be applied to the datasets and algorithms used in the sequel.

*4.1. Noise models and datasets*

We employ three types of image corruption models; i) additive white gaussian noise (AWGN), ii) impulse noise, and iii) speckle noise over 1000 randomly chosen images from the PASCAL dataset (Everingham, Van Gool, Williams, Winn, & Zisserman, 2010). All images are converted to grayscale and resized as 60x60 patches. We perform 10-fold cross-validation and compute the test performance as the average across all folds of the held-out test set. Specifically, in each fold, the network is trained on 100 patches, and tested on the remaining 900. In addition, we also evaluate the restoration performance of the trained model on noisy high-resolution grayscale images from BSD12 and BSD68 datasets.

In case of AWGN, we corrupt each of the images to have -5dB SNR (defined as the ratio of the image variance to the mean square error of the noise). For impulse noise denoising, a fixed-value impulse noise with a probability of 0.4 is applied; consequently replacing 40% of the pixels with either the darkest or brightest pixel values possible within the data range. For speckle noise, we employ the model used in (Bioucas-Dias & Figueiredo, 2010) where the noise probability is given by the Gamma distribution:

$$p(n) = \frac{M}{\Gamma(M)} e^{-nM} n^{M-1} \qquad (11)$$

For our experiments, corrupted images corresponding to acute noise level of $M=5$ are used. For all the problems, we minimize the mean-squared loss and use the Peak Signal-to-Noise Ratio (PSNR) as the performance metric:

$$PSNR(x_{orig}, x_{noisy}) = \frac{MAX_x^2}{\Sigma_N (x_{orig} - x_{noisy})^2} \qquad (12)$$

where $MAX_x$ is the maximum possible peak of $x$ in the data range.

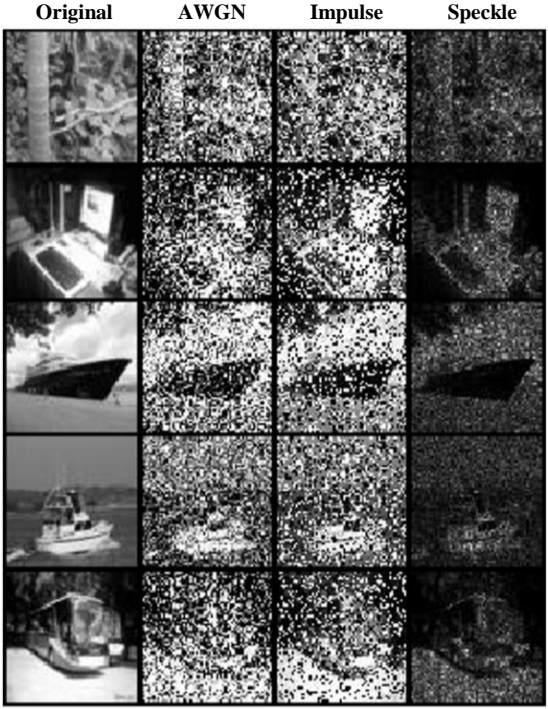

Figure 2. Examples of target patches and their corrupted counterparts for each type of noise.

*4.2. Network Architecture*

We use a compact network architecture comprising of 2 hidden layers, as depicted in Figure 3. Moreover, as presented in Section 3.2, a generative neuron has $Q$ times more network parameters compared to a convolutional neuron. Therefore, in order to investigate the exact impact of employing these additional learnable parameters, we conduct a series of ablation tests. Specifically, we compare a Self-ONN with the order $Q > 1$ to the corresponding CNN with the same number of neurons, as well as with the CNNs having roughly the same number of network parameters. This equivalence is reached by increasing the width of each convolutional layer until the number of network parameters are similar.

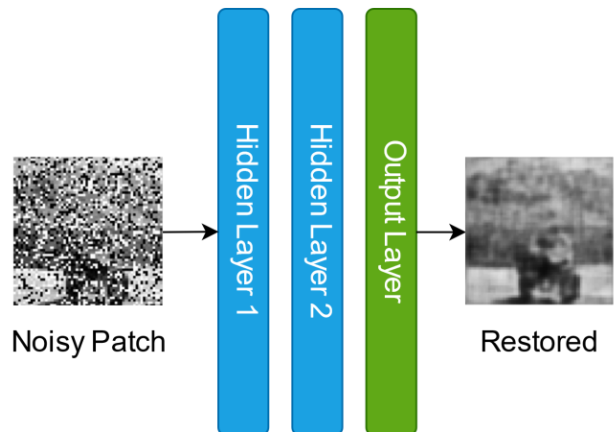

Figure 3. The network architecture employed in this study.

We propose three Self-ONN configurations; SelfONN-3, SelfONN-5 and SelfONN-7 corresponding to $Q \in [3, 5, 7]$, respectively. The number of generative neurons in the first and 2nd hidden layers are set to 6 and 10, respectively. The corresponding CNNs with the same number of parameters are termed CNN-3, CNN-5 and CNN-7, while the CNN having the same number of neurons as the Self-ONN is referred to as CNN-1 in the study. For Self-ONNs, with the use of $tanh$ as activation, the value of $a$ is naturally set to 0 in (8). Additionally, we restrict the choice of pooling function $\phi$ to summation. Table I provides the architectural details of each network. We also compare each Self-ONN to an equivalent ONN for AWGN. All the networks are trained using SGD with a learning rate of 0.01 and a momentum of 0.9, using 10-fold cross-validation on the training set.

Table I. Architectural details for the different networks.

| Network | Neurons in Layer 1 | Neurons in Layer 2 | Parameters (k) |
|---|---|---|---|
| SelfONN-7 | 6 | 10 | 26.3 |
| SelfONN-5 | 6 | 10 | 18.8 |
| SelfONN-3 | 6 | 10 | 11.3 |
| CNN-1 | 6 | 10 | 3.7 |
| CNN-3 | 11 | 18 | 11.2 |
| CNN-5 | 14 | 24 | 18.4 |
| CNN-7 | 18 | 27 | 26.3 |

## 5. Results and Discussion

Table III shows the restoration performance in terms of PSNR over the patches in the test set of Pascal, and over the high resolution images from BSD12 and BSD68 datasets for all the three restoration problems. Figure 6 depicts the training curves for the three problems and all the networks included in this study, while Figure 5 provides visual examples of the denoising performance of various networks.

### 5.1. Self-ONNs versus equivalent CNNs

We first aim to investigate the performance of generative neurons and convolutional neurons when treated as standalone learning units. To accomplish this, we compare Self-ONNs against the CNN with the same number of neurons (CNN-1) for all problems and datasets. We observe from Table III that Self-ONNs consistently achieve on average, 0.72dB, 2.95dB and 0.34dB PSNR improvement in the test set (generalization) performances across the AWGN, impulse and speckle noise models respectively. The performance gap is exceptionally high for the case of impulse noise across all three datasets. Moreover, even for the case of AWGN where the convolutional neurons are expected to fare better, we observe a clear gap in both training and generalization performances consistently across the three datasets. Furthermore, the performance gain with generative neurons seems to be correlated with the order, $Q$, as higher order Self-ONN (e.g., SelfONN-7 and SelfONN-5) achieves a better performance when compared to the lower order variant (e.g., SelfONN-3).

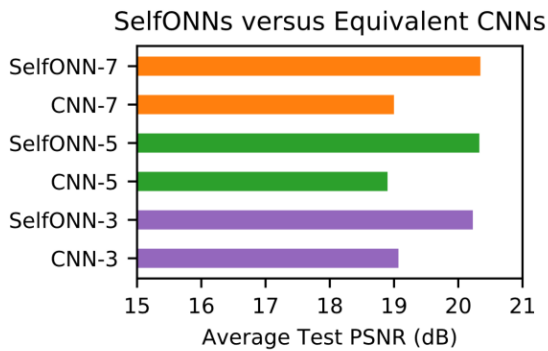

Figure 4. PSNR performances averaged across the three datasets for Self-ONNs with order $Q$ = 3, 5 and 7 as compared to equivalent CNNs over the test set. Colored bars with the same color correspond to networks with same number of network parameters.

Figure 4 shows the comparison of the generalization performance across all datasets and problems, between each Self-ONN configuration and its equivalent CNN, having the same number of network parameters. We can observe a clear trend where even a lower order Self-ONN such as the one with $Q = 3$, significantly outperforms its convolutional counterpart (more than 1dB PSNR improvement). This validates our claim that adding more parameters in the form of wider convolutional layers with more neurons is sub-optimal. In other words, instead of simply increasing the number of neurons, the generative neurons of Self-ONNs with optimized non-linear nodal operators achieve a better diversity and thus have a higher impact on the restoration performance.

*5.2. Self-ONNs versus equivalent ONNs*

In order to gauge the performance of Self-ONNs against equivalent size ONNs, we configured an ONN with the same network architecture as in Figure 3 using the operator combinations provided in (Kiranyaz et al., 2020) for AWGN noise. Table II provides the comparison over the test set of the resulting PSNR for Self-ONNs and ONNs across the three datasets.

Table II. Self-ONN and ONN PSNR performances on the restoration of images corrupted with AWGN.

|  | BSD12 | BSD68 | Pascal |
|---|---|---|---|
| **SelfONN-7** | **21.29** | **20.39** | **19.47** |
| **SelfONN-5** | 21.14 | 20.22 | 19.41 |
| **SelfONN-3** | 21.14 | 20.19 | 19.37 |
| **ONN** | 20.91 | 19.92 | 18.85 |

All Self-ONN variants outperform the ONN on this task across all three datasets. These results suggest that the nodal operators selected within the operator set library of ONNs, although better than the sole convolution operator of CNNs, may still not be optimal and the synthesized nodal operators in Self-ONNs provide a much better alternative.

Table III. PSNR performances for all models across the three noise types over the test sets.

| Noise | AWGN | | | IMPULSE | | | SPECKLE | | |
|---|---|---|---|---|---|---|---|---|---|
| **Network** | **BSD12** | **BSD68** | **Pascal** | **BSD12** | **BSD68** | **Pascal** | **BSD12** | **BSD68** | **Pascal** |
| **SelfONN-7** | **21.29** | **20.39** | **19.47** | **23.35** | 22.3 | 21.5 | 17.76 | 17.84 | 19.2 |
| **SelfONN-5** | 21.14 | 20.22 | 19.41 | 23.34 | **22.36** | **21.53** | **17.83** | **17.94** | 19.18 |
| **SelfONN-3** | 21.14 | 20.19 | 19.37 | 23.3 | 22.22 | 21.47 | 17.52 | 17.62 | **19.21** |
| **CNN-1** | 20.76 | 19.2 | 18.74 | 20.39 | 19.42 | 18.45 | 17.36 | 17.57 | 18.76 |
| **CNN-3** | 20.87 | 19.25 | 18.8 | 20.45 | 19.47 | 18.52 | 17.71 | 17.8 | 18.75 |
| **CNN-5** | 20.88 | 19.28 | 18.81 | 20.52 | 19.5 | 18.53 | 16.78 | 17.08 | 18.71 |
| **CNN-7** | 20.85 | 19.3 | 18.8 | 20.46 | 19.47 | 18.52 | 17.32 | 17.52 | 18.75 |

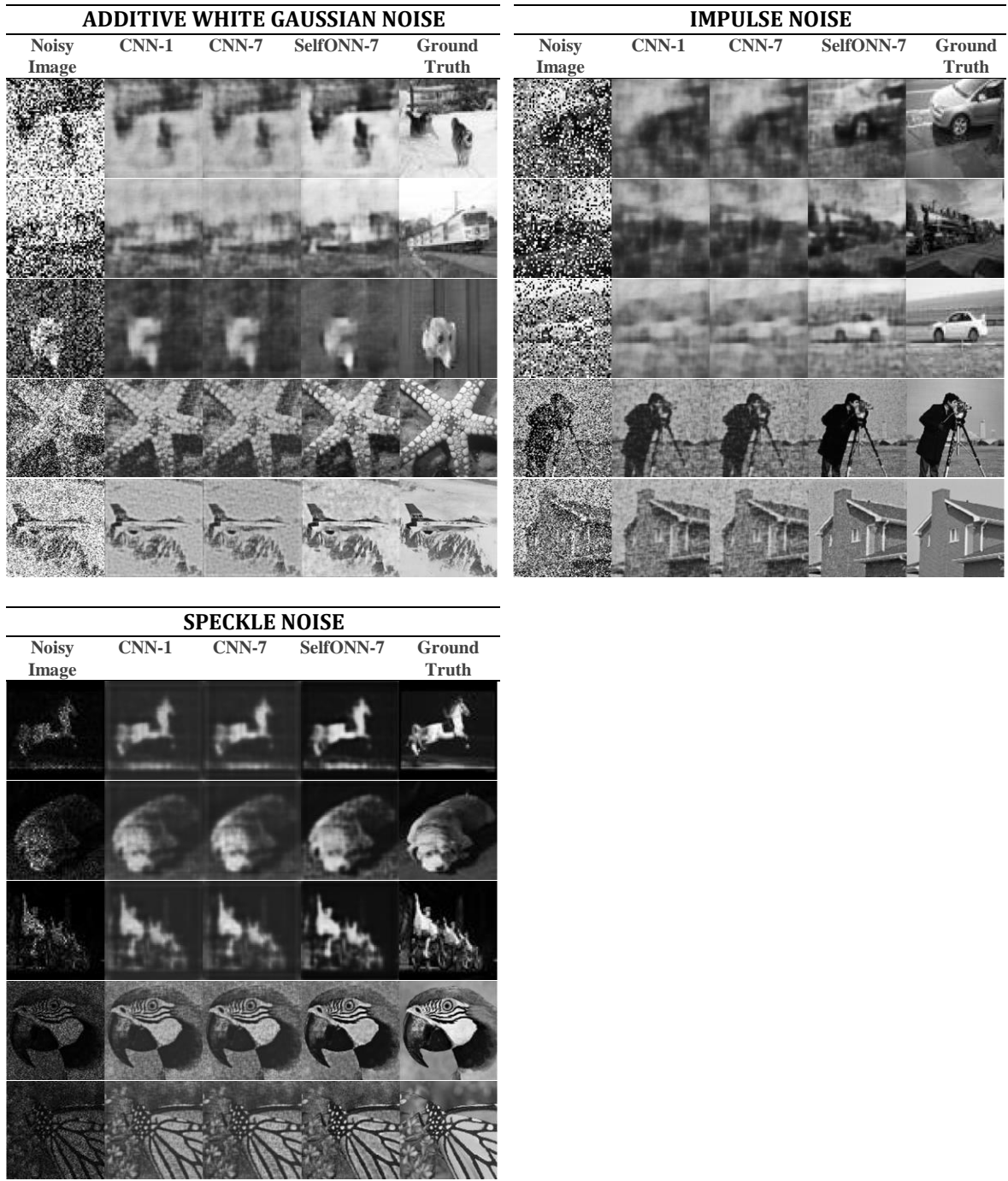

Figure 5. Sample restoration results from Pascal (patches) and BSD12 (high-resolution) datasets comparing the restoration performances.

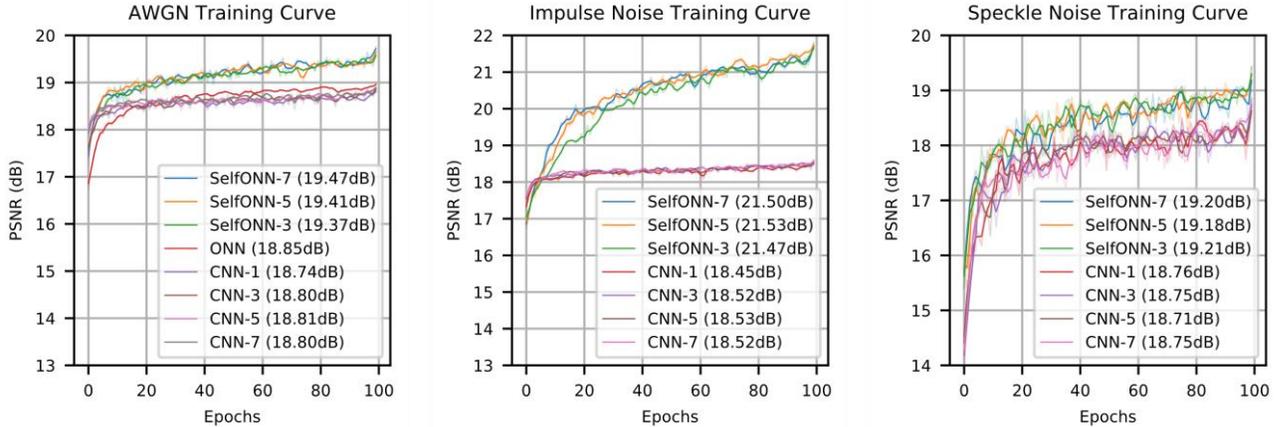

Figure 6. PSNR curves versus BP epochs for images restored from different noise models of all the networks.

## 5.3. Computational Complexity Analysis

Table IV provides the comparison of the total number of multiply accumulate (MAC) operations for the networks considered in this study. The number of MACs for the $l^{th}$ layer of the network is calculated using the following formula:

$$MACs(l) = |Y_l| * \left((N_{l-1} * K_x^l * K_y^l * Q^l) + 1\right)$$

where $|Y_l|$ is the number of elements in the output of the current layer, $N_{l-1}$ is the number of neurons in the previous layer, $K_x^l$ and $K_y^l$ are the kernel dimensions for the current layer, $Q^l$ is the order of approximation. The last term can be omitted for the special case where the bias is not used. In addition, for each network, we also calculate the ratio of MACs and the average generalization performance in terms of PSNR across all noise models and datasets. This provides a quantification of the utility of trainable parameters in each network architecture.

Table IV. Computational complexity of each network in terms of multiply-accumulate (MAC) operations.

|  | MACs (G) | MACs per dB (G/dB) |
| --- | --- | --- |
| **SelfONN-7** | 94.71 | 4.655 |
| **SelfONN-5** | 67.66 | 1.998 |
| **SelfONN-3** | 40.62 | 3.27 |
| **CNN-3** | 40.41 | 4.961 |
| **CNN-5** | 66.28 | 2.138 |
| **CNN-7** | 94.61 | 3.562 |

As expected, we see that Self-ONNs have roughly the same number of MACs as compared to their equivalent CNNs. However, the highest order Self-ONN, SelfONN-7, requires 4.65G MACs per dB of test PSNR, whereas the next lower order configuration, SelfONN-5, only consumes 1.99G flops per dB of generalization performance. There are two key insights which can be gained from this observation. Firstly, the difference implies that there exists a certain optimal value of $Q$ for a given problem for Self-ONNs, above which the generalization performance saturates for this particular problem. Secondly, there might exist a sparsity in the higher order weights, which can be exploited during inference to speed up performance using only a subset of the $Q$-dimensional weights, at the expense of negligible loss of performance. Both of these observations will be investigated further in our future studies.

## 6. Conclusions

In this study, we propose Self-ONNs to tackle severe image restoration problems. Self-ONNs are composed of generative neurons, which have the ability to synthesize any nodal operator by leveraging Taylor polynomials. Our results provide conclusive evidence that these optimized nodal transformations achieved, through generative neurons, considerably higher learning and generalization performance when compared with the linear mappings of convolutional neurons. Moreover, we show that adding learnable parameters by increasing the number of convolutional filters is sub-optimal, while doing so in a manner such that the added parameters can more directly influence the degree of non-linearity is a more worthy investment. The insufficient enrichment of the solution space when adding convolutional filters is one of the key reasons why state-of-the-art CNN architectures for image restoration are generally very deep. Finally, we show that the proposed Self-ONNs can even outperform its predecessor, ONNs, suggesting that for challenging inverse imaging problems such as the ones tackled in this study, hand-crafting an operator set library is not practical as the required nodal transformation may not exist in the form of well-known functions.

The core idea of generative neurons provides a modular interface and as such, it can be incorporated directly into the prevalent denoising and restoration architectures to increase their performance, as well as decrease the network size. Moreover, the networks used in this study were compact and used basic training paradigms such as SGD-based optimization. It will be an interesting research direction to explore various contemporary techniques for Self-ONNs essentially used for stabilizing the training of CNNs, such as batch normalization, Adam optimization with its variants, and dropout. These will be the subjects of our future research direction.